\documentclass[8pt]{article}
\usepackage[utf8]{inputenc}
\usepackage[utf8]{inputenc}
\usepackage[a4paper, margin=1.2in, top=0.7in]{geometry}
\usepackage{fancyhdr}
\usepackage{array}
\usepackage{pgfgantt}
\usepackage{graphicx}
\usepackage[export]{adjustbox}
\usepackage{ragged2e}
\usepackage{caption}
\usepackage{pdfpages}
\usepackage{float}
\usepackage{amsmath}
\usepackage{amssymb}
\usepackage{dsfont}
\usepackage[dvipsnames]{xcolor}
\usepackage[colorinlistoftodos]{todonotes}
\usepackage[style=ieee]{biblatex}
\title{Applying Language Models in Clinical Medicine: Recent Trends and Perspectives}
\author{Erik Aerts}
\date{}

\begin{document}

\maketitle

\section*{}
The use and applicability of artificial intelligence (AI) in medical research and clinical practice (I will mainly refer to "medical AI" and "medicine" throughout the text, but note that I intend to include the clinical aspect as well when making said references) is a fascinating topic. Whilst the need and potential for AI in medicine is generally accepted, the adoption or implementation of AI faces barriers in the field. This stems from both logistical standpoint with surrounding infrastructure and sensitivity of data, but also a more subjective standpoint from ones own ethical perspectives and attitudes towards the role of AI in society. Questions that often arise in the field are: \textit{to what extent should AI be used in medical practice?}, \textit{can AI bear responsibility when usage inevitably will cause someone harm?}, \textit{what performance from an AI is considered good enough?} and \textit{how can we trust the black-box nature of the system?}, just to name a few. Aligning the perspectives of experts from both the technical and medical fields is essential to one day implement and fully utilize useful AI for medicine. Nevertheless, this day may be some time down the road from where we are at the time of writing. This has not, however, hindered an active community from researching and developing AI across the many different fields of medicine. \\ 
\indent Over the last decade, deep learning (DL) has been at the forefront of development for AI in medicine. Tasks related to image analysis and prognostic prediction has seen popularity and recognition which has sparked active improvement and refinement over time. Novelty in medical DL is still being actively pursued indicated by publications in recent top-tier AI related conferences. Eye Gaze explores how information about where radiologists look at an image can be used to improve the connection between medical images and their accompanying reports \cite{Eyegaze}. The model uses a CLIP-based framework with multiple levels of learning. The images are divided into smaller patches and the reports into individual sentences, allowing the model to learn these more fine-grained relationships through contrastive learning. The model learns the behavioral patterns of radiologists of where in the images it can search to understand the highly domain specific language in the report. MOTOR focuses on a different problem highly relevant in the medical field: predicting time-to-event \cite{MOTOR}. Given a sequence of events in a patients medical history, a trained Transformer model embeds a representation of the patient trajectory and combines this with a survival model that divides time into intervals. For each interval, the model estimates the likelihood, or hazard, of an outcome occurring. Lastly, Medformer addresses the challenge of working with medical time-series data commonly found in medical practice as both monitoring and active assessment (ECGs, EEGs, EMGs etc) \cite{Medformer}. These types of medical time series often share properties such as multiple different views of the signal and how different granularity levels can affect the perspective of the patient status. The Medformer builds the architecture and training schema around these signal properties creating a model highly applicable to the data modality.\\
\indent The rise of large language models (LLMs) in the field of DL has been explosive. The idea of using a model primarily trained on language corpora to solve out of domain problems both with- and without external knowledge was previously unheard of. LLMs in theory solves a key restriction of traditional deep learning models: their inability to broaden their bandwidth of input/output dimension. In a medical context, this ability is highly sought after due to the unpredictability and flexibility needed with the available data to solve the at hand medical task. Back in 2022, the race to establish the superior LLM performance in medicine begun. A considerable threshold was passed when ChatGPT-3.5 managed to pass the USMLE (United States Medical License Examination) bar \cite{bolton2023foundation}. This was considered as a breakthrough, as the medical understanding needed by the LLM to pass such a bar was deemed to be at a high level at this point. Elliot Bolton explains in his talk how their model BioMedLM managed to reach a new SOTA performance on the MedQA benchmark weeks later, only to be broken again by MedPaLM 10 days afterwards. The main challenge was to find enough medical information available for training, and in his talk \textit{Training and Application of Language Models in Medicine} by Keno Bressen he explains how they used flashcards from medical students and wikipedia corpora to reach a consistent passrate on the USMLE bar benchmark \cite{bressem2023training}.\\
\indent Passing the USMLE bar of medicine is good progress, but it does not give sufficient evidence that LLMs are ready to solve medical tasks outside of the multichoice options provided in the exam. In his talk \textit{The Limited Impact of Medical Adaptation of LLMs and VLMs}, Daniel Jeong highlights a large flaw in the way that LLMs are benchmarked in medicine: The performance of medically tuned LLMs are often compared to other medically tuned LLMs or generic SOTA LLMs on medical tasks, but often not to their backbone LLM \cite{jeong2025limitedimpact}. In their work, they tested 10 medical LLMs versus their non medical counterpart in a zero-shot setting and found that the medical LLMs fail to consistently perform superior. Moreover, the medical corpora used for training data in medical LLMs can also be faulty as it can make LLMs perpetuate social biases in medical contexts. Kenza Benkirane explains in her talk \textit{How Can We Diagnose \& Treat Bias in LLMs for medical Decision-Making?} how introducing/removing both gender and ethnicity does play a role in neutral prompting for medical accuracy \cite{benkirane2025diagnose}. Some models heavily skew performance based on the tokens "men", "black" or "white" which is unwanted when providing safe and equal medical care to a universal audience. Furthermore, a as a callback to the talk by Keno Bressem, a large worry in using LLMs for medical care is hallucinations. Testing LLMs on benchmark datasets with adversarial samples for vulnerabilities could expose LLMs for suitability in medical environments and should be more researched. An interesting approach to generate adversarial examples was the DA3 method proposed by Wang et al which uses specific loss functions to generate adversarial examples more closely resembling real data \cite{bressem2023training, wang2024analyzing}.\\
\indent Based on the presented limitations, we can deduce that more high quality- and diverse benchmarks are needed to test developed LLMs in the multidimensional workflow clinicians do on a daily basis outside of standard multiclass classification. Recently, multiple interesting medical benchmarks has been proposed in a wide array of tasks associated with medical practice. PretexEval addresses the problem of outdated and potentially contaminated medical QA benchmarks by generating new evaluation samples from existing knowledge bases, MediConfusion focuses on multimodal reasoning, testing whether MLLMs can distinguish between visually similar radiology images and MedCalcBench instead evaluates whether LLMs can correctly perform medical calculations using calculator tools and synthetic patient cases \cite{PretexEval, MediConfusion, MedCalcBench}. Other benchmarks target broader aspects of medical practice. SM3-Text-to-Query evaluates whether LLMs can translate medical information into queries across different modalities and query languages and MedAgentBoard investigates whether multi-agent systems can improve performance across different medical tasks and domains, finding that collaboration does not consistently outperform monolithic or domain-specific models \cite{SM3TextQuery, MedAgentBoard}. Yang et al. proposed the PediatricsGPT accompanied with the PedCorpus dataset: a new benchmark and LLM tailored towards specifically pediatrics care made from pediatric textbooks, guidelines, and knowledge graph to construct specific multi-task queries \cite{PediatricsGPT}. To relate back to the concerns of clinicians that LLMs can be unsafe in practice, MedSafetyBench addresses this by testing whether LLMs recognize and refuse harmful medical requests. Using validated prompts by clinicians based on the AMA Principles of Medical Ethics, the benchmark found that many medical LLMs still comply with requests that could potentially cause harm \cite{MedSafetyBench}. This has been further expanded by Yin et al. with BingoGuard, a framework for recognizing and refusing to generate harmful content \cite{BingoGuard}. BingoGuard expands the interpretability of harm from a binary separation to different levels of security. These benchmarks illustrate a broader shift in LLM evaluation for medicine: the knowledge of medicine is shifted towards whether the models it can reason, use tools, interpret medical data and operate within healthcare infrastructure.\\
\indent Unlike the evaluation of the medical LLMs, the quality of the data corpora used to train the models has not seen as much attention in the literature. A lot of medical information is stored as free-text. This serves as a high potential source of information for models to train on. However, unlocking the full potential of these lakes of data requires a lot of precise work. In his talk \textit{Large Language Models as Universal Medical Forecasters}, Zeljko Kraljevic from University College London described this through their preparation work needed to homogenize available data to train on for forecasting patient trajectories \cite{kraljevic2024largelanguage}. Patient information is distributed across many different systems and formats, making it difficult to retrieve a complete patient history. Their introduced Foresight pipeline addresses in three steps: CogStack to harmonize and connect data from different systems, followed by MedCAT to extract and encode relevant biomedical concepts and AnonCAT anonymizes the resulting data. This allows them to construct longitudinal patient timelines, which were then used to train an LLM for forecasting patient trajectories. Similar challenges were explored by Hu et al. who prompted LLMs for named entity recognition (NER) in medical text \cite{hu2024improving}. This aimed to establish the utility of LLMs to extract useful terminology from EHR data, although the lack of comparison with traditional approaches such as MedCAT makes the practical benefit unclear. Parmar et al. approached the problem from another angle with BoX, a multitask BART model designed to handle different medical tasks through instruction prompting, including named entity recognition and question answering \cite{parmar2024instruction}.\\
\indent Another relevant topic in medical LLMs are resource allocation and requirements from an infrastructure point of view. Due to the sensitive nature of the data at hand, special attention is required when observing, sending and using patient information. Therefore, building local infrastructure is essential, often creating a constraint on the infrastructure from a cost and logistics perspective. With this in mind, developing resource effective training schemas and model architectures are of high importance in the field of medical AI. One interesting approach presented by Wang et al. is the FEDMEKI framework: a proposed framework to address the challenge of incorporating knowledge from multiple medical modalities without the need to share sensitive data to external sources \cite{FEDMEKI}. The framework trains models locally with onsite data across different modalities and subsequently aggregates their knowledge into a shared representation that can be used to guide the tuning of a foundation model on publicly available data. This provides a way of fusing general medical knowledge while keeping the underlying patient data local. In contrast, the talk \textit{LoRKD: Low-Rank Knowledge Decomposition for Medical Foundation Models} by Yao approaches the problem from the opposite direction: \textit{what if we already had a capable medical foundation model, but could not deploy it due to its resource requirements?} \cite{yao2025lorkd}. Their proposed LORKD framework distills knowledge from a large foundation model into smaller expert modules which can be deployed easier in resource scarce settings. This approach assumes that we have access to a good enough foundation model to distill data to which is something we have observed to be a bit out of reach at the time of writing due to difficult in the medical corpora used to train the models on. Interestingly, Zhang et al. argue that LLMs can outperform well in a medical setting but their inclusion in a medical context is purely resource based \cite{zhang2024enhancingsmallmedicallearners}. They use this reasoning to argue for their pipeline to extract knowledge from LLMs to train smaller models which are easier to host locally comply with patient confidentiality. Whilst the idea of implementing locally hosted smaller models with domain specificity remains an appropriate choice, the claim of LLMs being "medically capable" can be put into question as it relies on the USMLE benchmark being enough to establish medical knowledge which one might argue with based on previous discussions in this post. Lastly, the SERSAL framework by Yan et al. addressees the challenge of LLMs and working with tabular data; something very relevant in medical data sets of which the tabular modality is very common \cite{SERSAL}. Instead of relying full scale pre-training for LLMs which is high cost for a resource limited environment, the SERSAL framework creates a synergetic learning process between an LLM and a smaller specialized model for tabular prediction. The learning scheme uses levels of uncertainty to create soft pseudo labels from zero-shot prompting of LLMs to teach the smaller supervised model to adapt the pseudo labels into a supervised learning scheme. The smaller model then relates the feedback to the LLM to revise small aspects of the LLM tuning, creating an iterative cycle to improve LLM reasoning with tabular data. \\ \\
As described throughout the post, LLMs have not shown sufficient parametric medical knowledge to reach a satisfactory level yet. The field has given a lot of attention to creating diverse benchmarks, making good use of the available data and how to solve potential resource restrictions in the field. Together, these efforts highlight the clear limitations of the current state of the art and the challenges that remain before LLMs can eventually be deployed in medical care. For this, the ability for the LLM to gain domain specific knowledge and interpretations from external sources is pivotal in the development of medical LLMs. One such field is toolcalling for LLMs: a field which has been given a lot of attention in recent literature. Foundational approaches such as Toolformer and ReAct demonstrated the potential for LLMs to interact with external tools and incorporate their outputs into a broader reasoning process. The interaction between LLM and the tool environment have seen further development in recent literature \cite{Toolformer2023, Yao2023React}. MetaTOOL provided a useful categorization of the four stages of toolcalling: (1) \textbf{decision} of whether the LLM needs a tool for the given query, (2) \textbf{identification} of which tool(s) to use for the given query, (3) \textbf{call} from the LLM to the selected tool(s) and, (4) \textbf{interpretation} of the tool(s) feedback to the LLM to use for reasoning \cite{MetaTool}. Huang et al. argue that a large focus has been placed on stages (3) and (4), and hence their proposed MetaTOOL benchmark dataset give more attention to stages (1) and (2) in the toolcalling process. The authors found that most state-of-the-art LLMs for toolcalling lack standout performance in the stages (1) and (2) motivating more attention.\\
\indent The limitations identified by MetaTOOL can be found as motivating undertones in recent literature. CRAFT, a framework proposed by by Yuan et al., combined tool synthesis and retrieval for domain-specific tool use. CRAFT first generates tools from problems sampled from a dataset by generating Python solutions, abstracting them into more general purpose functions \cite{CRAFT}. This creates a toolbox tailored to a specific domain. For tool retrieval and identification, CRAFT introduces a multi-view retrieval process matching: (1) the current problem to the original problem prompting the formulation of the tool, (2) the predicted function name of what is needed to the actual tool name and, (3) the predicted docstring of what is needed to solve a task to the tool documentation. This aims to improve the identification of the appropriate tool for a given query. Continuing, Chen et al. introduce the ToolEVO framework which treats tool identification as a dynamic process rather than assuming a static tool environment \cite{ToolEVO}. Using Monte Carlo Tree Search, the framework can select, invoke, and update tools based on environmental feedback allowing the LLM to adapt its tool selection when tools or APIs change. Liu et al. instead approach the field from the perspective of tool failure during planning \cite{ToolPlanner}. Their framework Tool-Planner clusters tools into toolboxes based on automatically generated functionality labels and text embeddings allowing the LLM to select a functionality rather than committing to a specific API. If a selected tool fails, another tool from the same toolbox can therefore be substituted without recalculating the entire plan. Together, these works highlight that tool identification extends beyond simply selecting an API: effective identification can require domain specific retrieval, adaptation to changing tool environments, and the ability to substitute tools when the initially selected option is unavailable.\\
\indent All the provided examples from recent literature has been single turn question-answering pairs with toolcalling. However, in many environments of deployments, the interactions will be multiturn question-answering with feedback from a user. Wang et al. addressed this with the proposal of the MINT benchmark for LLMs performance during multiturn interaction focusing on tool augmented task solving and natural language feedback \cite{MINT}. The authors found a few distinctions through their experiments: (1) LLM benefits more from tools than they do from language feedback for a given task, (2) better single turn performance does not translate to better multiturn performance (note: specifically for solving tasks, not for picking the right tool for a task) and, (3) reinforcement learning from human feedback harms multiturn interactions between LLM and tools. What is noteworthy for the mentioned methodology of Wang et al. is that the prompting was set up to encourage the model to actively call tools and let it know that there will be multiple turns of questions. This may introduce some bias in the evaluation of the capability of the LLM with tools, specifically regarding the undervalued categories for awareness and tool selection highlighted by MetaTOOL. 

\section*{Individual reflections}
When working with medical AI, a problematic area of the field that often slows down progress is benchmarking the results. The tasks to solve in the field has progressively gotten more narrow meaning that the specific questions different teams are trying to answer are not compatible. Be it a specific comorbidity, medical setting or timeframe the cohort at hand for a given research question can be incompatible with previous cohorts. A few years back, Bolton talked about the gold standard benchmarking of medical AI being the bar exam for medical licenses in his talk. The early benchmarks solely circulated around pinpointing medical knowledge and how much corpora the LLM understood, such as the flash cards for medical students used by Bressem et al. This is a far stretch from what we actually want medical AI to do in practice, prompting the developed in many different directions, and introducing benchmarks for medical AI in diverse tasks is currently considered novel in many venues. Circling back to the opening problem formulation, it feels like the field is in a transitional era in regards to benchmarking. We have recognized the flaws of the initial benchmarking which only tested medical knowledge from a limited view of recognizing terms to attempting to nail down more specifically what we want the LLMs to be capable of doing in a medical setting. However, we are still missing the connection between LLM benchmarking and specific domains within the medical infrastructure as practice often differs between the settings. A notable standout from the crowd of benchmarks discussed in this text is the PedCorpus by Yang et al. The aim of creating a benchmark specifically for LLM usage in pediatrics creates a more clear medical subdomain. If LLM are aimed to be used in pediatric care, the benchmark is highly specific for the aforementioned field mapping the progress along the way. \\
\indent Over time, I believe the field will be moving towards primarily handling medical free text. The talk by Kraljevic highlights well how available data is formatted and stored from a medical perspective. The tailored components in the Foresight pipeline with techniques such as NER shows specifically the tasks we need to solve in order to utilize the free text data spread out over multiple systems. This aligns with the other talks where the presenters either work alongside clinicians, or are clinicians themselves, where a clear focus is shown towards making LLMs useful from a resource allocated perspective helping clinicians understand the vast corpora of medical information. This idea of medical AI using the backend information is not shared with the common tasks published in the newer benchmarks. A clear focus on question-answering pairs is still present which might not be the most viable test of LLMs capability in a medical setting. Firstly, as discussed previously, the need and environment in medicine requires assistance and use of other queries. Secondly, The type of questions clinicians would ask has not been fully mapped out as more domain specificity in the query or longer conversations are not usually present in the benchmarks. Thirdly, it has been observed that the medical foundation models might not be fully ready for medical tasks as the general understanding is that the parametric knowledge of medical foundation models are limited. The talk by Jeong highlighted this well with the limited impact medical finetuning showed in their experiments. Moreover, a recent study by Bean et al. suggested that LLMs are not yet ready for deployment as standalone medical assistants in direct patient care affirming the view on what type of role LLMs may be best suited towards at the moment \cite{Bean2026}. \\ 
\indent An interesting trend was found regarding utilizing models of different sizes to find effective training schemes for both foundation models and LLMs using a combined effort to transfer medical knowledge. This topic is highly relevant from a medical context as resource restrictions needs to be taken into consideration when building systems or solutions aimed to be useful in medical care. Moreover, as mentioned previously, the sensitivity of the available data also puts restrictions on which type of model can be used based on the hosting. If the hosting is done externally by a company the regulations and laws may restrict usage leading to a importance of being able to locally host models to use. Frameworks such as SERSAL, FEDMEKI and LorKD are highly valuable when trying to tune larger models with either limited computational resources or limited data to train on. However, some of the assumptions made by the frameworks should be in consideration before applying the frameworks directly. \\
\indent  Toolcalling has been popular in recent literature and has shown progress in creating benchmarks for the LLMs ability to interact with tool environments which may seem more tailored towards domain specific purposes. However, their direct applicability to the medical domain can be brought into question. An example of this is the CRAFT framework by Yuan et al.: the idea of LLM synthesizing its own tools to solve perceived tasks can be unreliable in fields such as prognostic medicine and diagnostics where medical tools often are based on observed trends in medical cohorts. However the calling framework of CRAFT based on different views can be a foundation for medical toolcalling. When calling tools, different medical views should be accounted for such as patient aligning with the development cohort, requested outcome compared to what the medical tool gives and reliability of the tool itself. The dynamic interpretation of the tool environment presented in ToolEVO prompting the Monte Carlo Tree Search pathing also has applications in a medical settings. Over time, more tools are developed from up-to-date guidelines which may see older versions be redundant or less accurate. However, the progress rate of the evolving nature of medical tool environment compared to a generic company may be starkly different, and the implementation of ToolEVO might not be needed to solve what is most likely a easier problem in a medical setting than the authors claim. Similarly, the Tool-Planner also addresses a valid concern in general purpose toolcalling which has correlations with medical care but not to the extent that the authors intend. Clustering tools based on medical field, outcome, comorbidity and so on makes sense in theory. However, what can affect applicability of such an implementation from a medical setting would be the bias in the data based on the setting. Clustering tools related to heart failure is a valid approach, however if a tool fails to be invoked due to reasons related to missing information for example, the idea of invoking a similar tool from the same toolbox would likely encounter a similar fate. medical tools in a given field often have related features with high relevance for the specific outcome in question. If a key feature is missing from one tool causing a failed invocation, replanning a new invocation from a similar tool may likely face the same bottleneck. Overall, general trends found in general purpose toolcalling can find applicability in medical care. What should be noted is how the domain can change the perception of these general purpose problems to either complicate the initial proposition or in other cases like ToolEVO possibly simplify the initial suggestion.

\printbibliography

\end{document}